\documentclass{article}

\usepackage[final]{neurips_2025}

\usepackage[utf8]{inputenc} 
\usepackage[T1]{fontenc}    
\usepackage{url}            
\usepackage{booktabs}       
\usepackage{amsfonts}       
\usepackage{graphicx}       
\usepackage{nicefrac}       
\usepackage{microtype}      
\usepackage{xcolor}         
\usepackage{amsmath}
\usepackage{multirow}
\usepackage{multicol}
\usepackage{tcolorbox}
\usepackage{subfigure}
\usepackage{makecell}
\usepackage{placeins}
\usepackage{color, colortbl}
\definecolor{citecolor}{HTML}{A5C09F}
\usepackage[breaklinks=true,bookmarks=false, colorlinks=true, linkcolor=ForestGreen,bookmarks=false, citecolor=citecolor]{hyperref}

\usepackage{natbib}

\usepackage{tcolorbox}
\usepackage{placeins}
\tcbset{
  myboxstyle/.style={
    colback=gray!5,      
    colframe=gray!80,    
    boxrule=0.5pt,       
    arc=4pt,             
    outer arc=4pt,
    boxsep=5pt,          
    left=5pt,            
    right=5pt,           
    top=5pt,             
    bottom=5pt,          
  }
}
\usepackage{caption}

\definecolor{light-gray}{gray}{0.6}
\definecolor{front-color}{HTML}{F5FFFA}
\definecolor{Gray}{gray}{0.93}

\newcommand{\best}[1]{\textbf{\underline{#1}}} 
\newcommand{\sbest}[1]{\textbf{#1}}    
 
\definecolor{ForestGreen}{rgb}{0.13, 0.55, 0.13}
\newcommand{\methodname}{Visual Evidence Reward}
\newcommand{\modelname}{Video-VER}
\definecolor{LighterGray}{gray}{0.93}
\definecolor{LightPurple}{RGB}{232,244,234}

\newcommand{\thickhline}{\Xhline{3\arrayrulewidth}}

\newcommand{\custompar}[1]{
  \par
  \vspace{2pt}
  \noindent\textbf{#1}
}

\newcommand{\startexcludefrom}[1]{
    \addtocontents{toc}{\protect\setcounter{tocdepth}{-1}} 
    #1
    \addtocontents{toc}{\protect\setcounter{tocdepth}{3}} 
}

\title{
 \raisebox{-0.18cm}
 {\includegraphics[scale=.035]{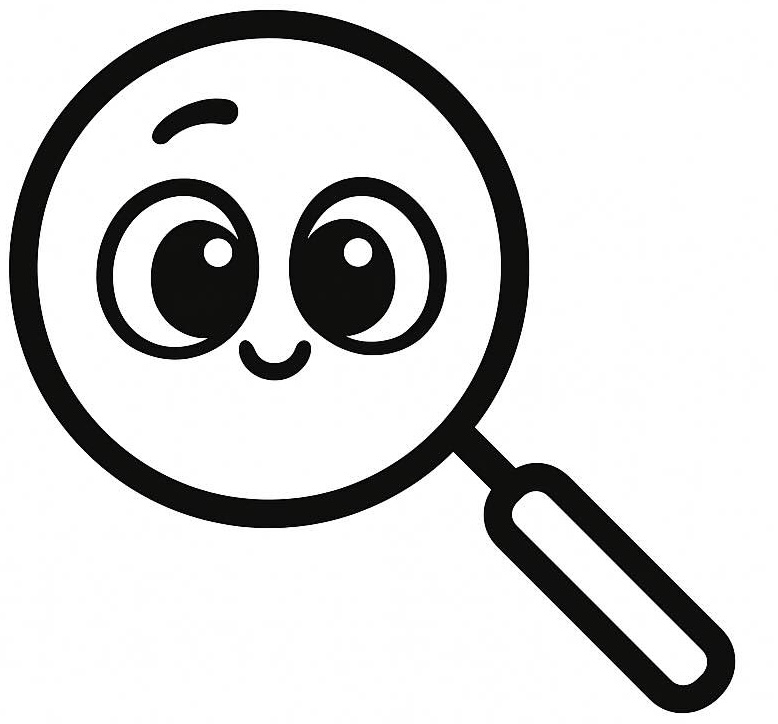}}
When Thinking Drifts: \\ Evidential Grounding for Robust Video Reasoning}

\author{
Mi Luo\textsuperscript{1}\space\space\space 
Zihui Xue\textsuperscript{1}\space\space\space 
Alex Dimakis\textsuperscript{2, 3}\space\space\space 
Kristen Grauman\textsuperscript{1}\space\space\space
\vspace{1mm} \\
\textsuperscript{1}The University of Texas at Austin\qquad 
\textsuperscript{2}UC Berkeley\qquad 
\textsuperscript{3}Bespoke Labs \vspace{1mm}\\
}

\begin{document}

\maketitle

\startexcludefrom{
\begin{abstract}
Video reasoning, the task of enabling machines to infer from dynamic visual content through multi-step logic, is crucial for advanced AI. While the Chain-of-Thought (CoT) mechanism has enhanced reasoning in text-based tasks, its application to video understanding remains underexplored. This paper presents a systematic analysis revealing that CoT often degrades performance in video reasoning, generating verbose but misleading internal monologues, and leading to hallucinated visual details and overridden correct intuitions—a phenomenon we term "visual thinking drift." We explain this drift through a Bayesian lens, positing that CoT traces often diverge from actual visual evidence, instead amplifying internal biases or language priors, causing models to storytell rather than engage in grounded reasoning. To counteract this, we introduce \methodname~(VER), a novel reinforcement learning framework that explicitly rewards the generation of reasoning traces that are verifiably grounded in visual evidence. Comprehensive evaluation across 10 diverse video understanding benchmarks demonstrates that our \modelname~consistently achieves top performance. 
Our work sheds light on the distinct challenges of video-centric reasoning and encourages the development of AI that robustly grounds its inferences in visual evidence---for large multimodal models that not only ``think before answering", but also ``see while thinking".
\end{abstract}
\section{Introduction}
Imagine watching a cooking tutorial video: As the chef chops vegetables, combines ingredients, and adjusts the heat, we are not just passively observing isolated actions. Our minds actively connect these steps, anticipating the next move, understanding the purpose behind each technique, and envisioning the delicious outcome. This  act of deriving understanding from a sequence of visual events, inferring intent, and predicting results through a series of logical thoughts, mirrors the core challenge addressed by \emph{video reasoning}: empowering machines to draw inferences and conclusions from dynamic content of video through multi-step logical reasoning~\cite{fei2024videoofthought, mao2022dynamic,nagrani2025minerva, feng2025videor1, wang2024videoagent, min2024morevqa}.

As researchers strive to imbue systems with this capability, their efforts have largely branched into two complementary strategies. One prominent path is the agentic workflow, also referred to as modular reasoning~\cite{min2024morevqa}, which deconstructs video understanding into focused sub-problems—such as event parsing, object tracking, and temporal grounding addressed by specialized models~\cite{yu2023self, fan2024videoagent, wang2024videoagent, liu2025videomind, min2024morevqa}. 
Alongside this modular approach, a parallel evolution in Multimodal Large Language Models (MLLMs) has ushered in a more holistic reasoning paradigm, where a model executes end-to-end inference by 
harnessing both vision and text~\cite{Qwen2.5-VL, li2024llava, zhang2024llavanext-video, zhang2023videollama, zhang2024longva, cho2025PerceptionLM, lin2023vila, li2023videochat, chen2024internvl}. Within this architecture, the vision encoder typically assumes the role of a perceptual engine, tasked with extracting high-fidelity visual representations, whereas  the language model component takes the helm of the reasoning process itself, generating textual rationales and constructing logical chains.

\begin{figure}[t]
    \centering
    \includegraphics[width=1.0\textwidth]{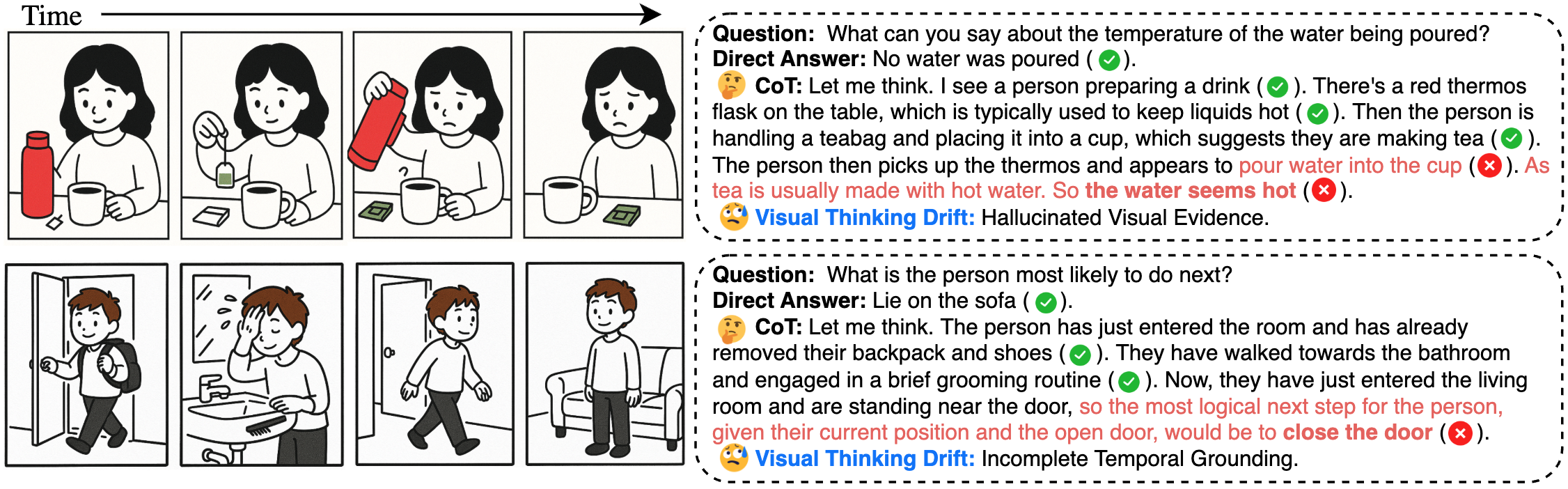}
    \caption{Two examples of \emph{Visual Thinking Drift} phenomenon, where the reasoning chain, as it grows longer, increasingly relies on hallucinated facts or incomplete temporal context—drawing conclusions from language patterns rather than grounding in the actual video content.}
    \label{fig:visual_thinking_drift}
\end{figure}

\newpage
Building upon this foundational capability, a significant body of recent work explores chain-of-thought (CoT) reasoning for MLLMs, both through high-quality reasoning datasets with spatio-temporal annotations~\cite{shi2024aotd, lu2025vited, han2024videoespresso} as well as reinforcement learning (RL) post-training approaches inspired by influential image- and text-based reasoners~\cite{guo2025deepseek,hu2025open,skywork2025r1v,team2025kimi,wei2022chain} that refine the MLLM's reasoning pathways~\cite{feng2025videor1, zhang2025tinyllava, li2025videochatr1} with rule-based rewards.  Early results suggest that simply encouraging ``thinking before answering" can often pay off.
However, despite these promising advancements, the unique challenges encountered when transposing \emph{text}-based chain-of-thought reasoning to the distinct demands of \emph{video}-centric tasks warrant deeper exploration.

In this paper, we aim to both expose the gaps in CoT-based video reasoning, and propose a solution.  First we present a systematic study covering 10 video benchmarks, multiple MLLMs, and 20 video QA subtasks, revealing that 
CoT reasoning often backfires for video understanding, especially with open-sourced models. 
We identify a recurring failure mode we term “\emph{Visual Thinking Drift}” where an MLLM introduces hallucinated facts or bias to outdated frames for temporal reasoning.  
Rather than enhance reasoning, the CoT prompts frequently induce models to produce verbose but misleading internal monologues—hallucinating visual details, overriding correct instincts, and ultimately reducing accuracy.  See Figure~\ref{fig:visual_thinking_drift}. For instance, in next-action prediction tasks, the model may base its logic on earlier events while ignoring more recent cues—despite being able to answer correctly when prompted directly, without CoT.  This drift reveals a critical flaw: CoT reasoning in video models often becomes performative rather than grounded—fluent, plausible, but ultimately wrong.  To understand this phenomenon, we adopt a Bayesian lens, showing that CoT traces often diverge from actual visual evidence and instead amplify internal biases or language priors.  

Next, to  counter the visual thinking drift  problem, we introduce \methodname~(VER), a novel reward mechanism for reinforcement learning-based MLLM post-training framework. Our VER explicitly encourages reasoning traces grounded in visual evidence. The key insight is that genuine video reasoning emerges when the internal thought process itself is actively and granularly tethered to perceived content, compelling models to truly ``see while thinking", not just ``think before answering". In the proposed model, an auxiliary LLM acts as a judge, evaluating the factual alignment between intermediate thoughts and visual inputs. This automatically encourages not just coherent but correct reasoning, stabilizing inference, and boosting overall accuracy.

We evaluate our \modelname~model across 10 diverse video understanding benchmarks. Compared to strong base models and existing reasoning techniques, \modelname~consistently ranks first or second. 
Furthermore, our model achieves consistently strong margins compared to its respective base MLLM (trained without the \methodname)---as much as +9.0\% absolute accuracy gains, and an average of +4.0\% across all 10 benchmarks.
Our results suggest that in video reasoning, grounding—not verbosity—is essential to true video intelligence.

\section{Related Work}
\custompar{Eliciting Reasoning Ability from Large Language Models}
Large language models (LLMs) have shown strong performance on complex reasoning tasks such as mathematics and programming~\cite{brown2020fewshot, chowdhery2023palm, team2023gemini, achiam2023gpt4, shao2024deepseekmath}. These capabilities are often elicited through few-shot~\cite{wei2022chain, chen2022program, yao2023tree, zhou2022least} and zero-shot prompting~\cite{yasunaga2023large, kojima2022large}, or through instruction tuning with large-scale chain-of-thought (CoT) datasets~\cite{chung2024scaling, nye2021show, OpenThoughts}. Recent advances show that even simple rule-based incentive mechanisms and lightweight reinforcement learning can induce robust reasoning without explicit supervision~\cite{guo2025deepseek}. However, studies also highlight that LLM-generated reasoning traces may be unreliable or unfaithful to the model’s internal processes~\cite{paul2024making, turpin2023language}. Motivated by these insights, we investigate how to elicit reasoning abilities from multimodal LLMs for video understanding---a domain that introduces unique challenges due to the dynamic temporal nature of video data.

\custompar{Video Reasoning}  Video reasoning entails drawing conclusions from video content through multi-step logical inference~\cite{fei2024videoofthought}. As overviewed above, research in this area follows two directions: modular reasoning~\cite{min2024morevqa} that decomposes tasks into addressable subcomponents~\cite{yu2023self, fan2024videoagent, wang2024videoagent, liu2025videomind, min2024morevqa} and MLLMs that perform end-to-end inference by jointly leveraging visual and textual information~\cite{Qwen2.5-VL, li2024llava, zhang2024llavanext-video, zhang2023videollama, zhang2024longva, cho2025PerceptionLM, lin2023vila, li2023videochat, chen2024internvl}. 

Building on this, some recent work explores enhanced chain-of-thought (CoT) reasoning for MLLMs. One line focuses on constructing high-quality reasoning datasets, grounded temporally or spatially, to guide more structured logic generation~\cite{shi2024aotd, lu2025vited, han2024videoespresso}. Another emerging direction, inspired by DeepSeek-R1~\cite{guo2025deepseek}, Open Reasoner Zero~\cite{hu2025open}, Skywork R1V~\cite{skywork2025r1v}  and Kimi k1.5~\cite{team2025kimi}, applies reinforcement learning (RL)~\cite{sutton1999reinforcement, luong2024reft} to refine the reasoning process through lightweight, targeted reward mechanisms~\cite{feng2025videor1, zhang2025tinyllava, li2025videochatr1}.
Despite these promising developments, systematic analyses of challenges of text-based chain-of-thought reasoning in video tasks remain limited. We contribute to this space by offering both empirical insights and a simple yet effective reward strategy designed to improve the faithfulness to visual content for reasoning chains. 

\custompar{Hallucination in MLLMs}
We identify a novel phenomenon termed "\emph{Visual Thinking Drift}", a specific manifestation of hallucination in MLLMs. While hallucination—producing descriptions, or conclusions misaligned with visual input—has long been a persistent challenge for MLLMs~\cite{bai2024hallucinationsurvey}, visual thinking drift is distinguished by its emergence within chain-of-thought reasoning: errors introduced at earlier reasoning steps, once hallucinated, can propagate through subsequent steps, ultimately leading to conclusions that significantly diverge from the visual evidence. In the image domain, prior research has primarily focused on object hallucination, where models misidentify object categories, attributes, or relationships~\cite{bai2024hallucinationsurvey}. In the video domain, hallucinations involve misinterpretations of dynamic actions, events, and narrative sequences~\cite{videohallucer, eventhallusion}. To mitigate such issues, existing work explores test-time interventions~\cite{liu2024steering, wang2024mllm, huo2024self, li2025IMCCD} and preference modeling during training to reduce vision-language misalignment~\cite{xie2024vdpo, zhang2024llavahound}.
In contrast, we propose a lightweight and data-efficient alternative: reinforcement fine-tuning to mitigate hallucination within the reasoning process itself. Rather than focusing solely on perception-level correction, our approach targets the integrity of logical progression, aiming to curb the cascading effects of visual thinking drift.

\section{Dilemma of Chain-of-Thought Reasoning in Video Understanding}
The standard approach to evaluating Video LLMs (a.k.a., MLLMs) for Video Question Answering (VQA)~\cite{li2024mvbench, fu2024videomme} focuses on their ability to provide direct answers—such as returning an integer for a question like ``How many objects enter the scene?'' 
Recent advances in LLM reasoning, particularly Chain-of-Thought (CoT) prompting~\cite{wei2022chain}, have encouraged models to reason step by step. This approach offers benefits in decomposing complex questions and improving the explainability of responses. In this study, we aim to systematically evaluate the two core prompting strategies: \textit{Direct Response Generation} and \textit{Reasoning-Driven Generation (Chain-of-Thought)}.
\custompar{Direct Response Generation} Under this scheme, the video LLM produces the output answer \( a \) by directly leveraging the input question \( q \) and the accompanying video context \( \mathbf{v} \). The generation task can be formally expressed as:
\[
p(a \mid q, \mathbf{v}).
\]
In this approach, the model is expected to yield the final answer immediately, without constructing any intermediate reasoning path or justification.  
\custompar{Reasoning-Driven Generation (Chain-of-Thought)}  
Conversely, this method decomposes the output generation into two sequential phases. Initially, the model infers a rationale sequence \( c_{1:T} \) based on the input query \( q \) and the contextual video data \( \mathbf{v} \). Subsequently, it conditions the final prediction \( a \) on both the generated rationale and the original inputs. This approach is captured by the following formulation:
\[
p(c_{1:T} \mid q, \mathbf{v}) \cdot p(a \mid c_{1:T}, q, \mathbf{v}).
\]

\begin{figure}[t!]
    \centering
    \includegraphics[width=0.99\textwidth]{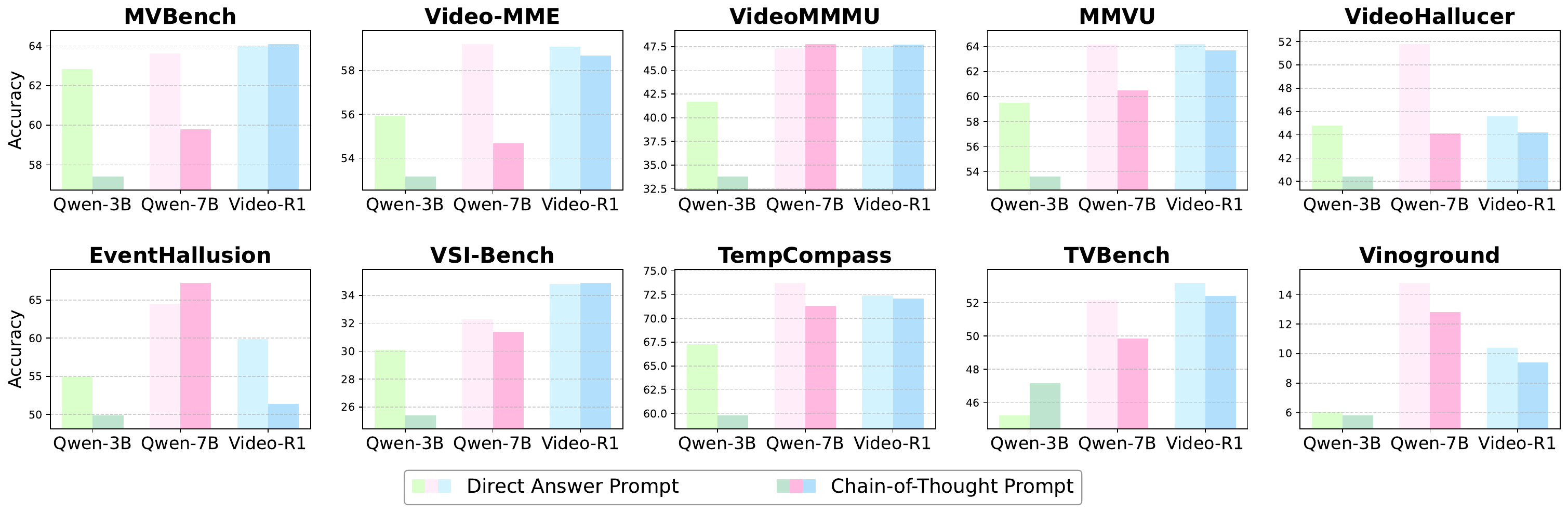}
    \caption{Compared to directly prompting the model for an answer, instructing the model to "think before answering" leads to a noticeable performance drop in open-source MLLMs such as Qwen2.5-VL~\cite{Qwen2.5-VL} and Video-R1~\cite{feng2025videor1} across multiple benchmarks (10 are shown here).}
    \label{fig:bar_cot_harm}
\end{figure}

\begin{figure}[t]
    \centering
    \includegraphics[width=0.99\textwidth]{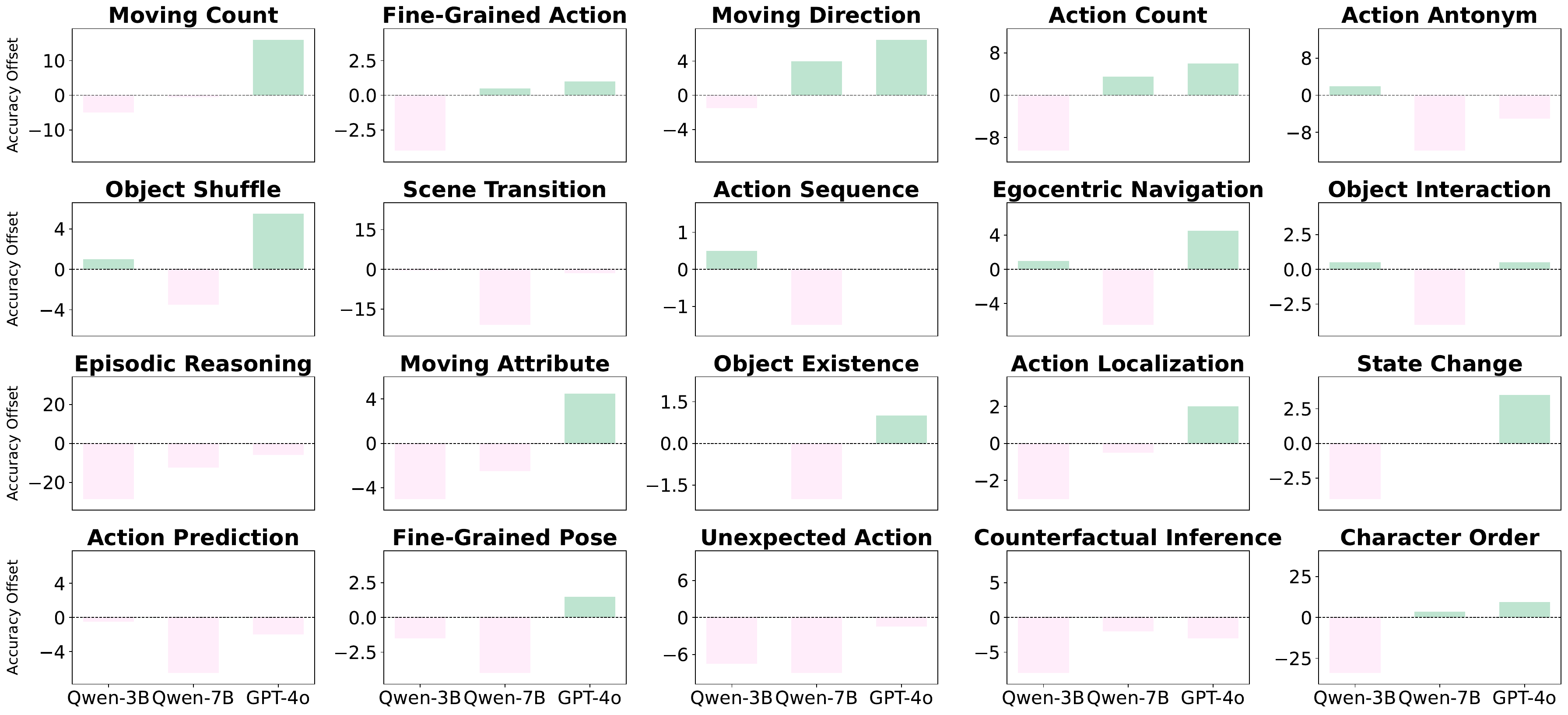}
    \caption{Gains (green) and losses (pink) with CoT prompt, showing that reasoning-driven generation is valuable for multi-hop, causal, or interpretability-driven tasks like object counting, but weakens both large and small models on lightweight perceptual questions such as scene transition detection.}
    \label{fig:mvbench_subtask_acc}
\end{figure}

\subsection{How Does CoT Make Models Weaker on Simple Video Perception Tasks?} 
To assess the impact of reasoning-driven generation on video understanding, we conduct a systematic study comparing it to direct response generation. Our key question: Does prompting models to "think step-by-step" improve performance in state-of-the-art Video LLMs?

As illustrated in Figure~\ref{fig:bar_cot_harm}, we evaluate both generation strategies across three leading open-sourced MLLMs—Qwen2.5-VL-3B~\cite{Qwen2.5-VL}, Qwen2.5-VL-7B~\cite{Qwen2.5-VL}, and Video-R1-7B~\cite{feng2025videor1}—on 10 diverse video benchmarks, spanning general video understanding~\cite{fu2024videomme} to complex temporal~\cite{zhang2024vinoground} and spatial reasoning tasks~\cite{yang2024vsibench}.\footnote{We also experimented with LLaVa-OneVision-7B~\cite{li2024llava}, but it failed to follow the instruction to generate a thought trace, likely due to its use of a weaker underlying language model.} Surprisingly, reasoning-driven (CoT) prompting often leads to lower accuracy, particularly on benchmarks demanding rapid visual perception, like Video-MME~\cite{fu2024videomme}.

To further dissect the impact of CoT prompting, we analyze 20 subtasks from MVBench (Figure~\ref{fig:mvbench_subtask_acc}), leveraging its structured task taxonomy. The results show that forcing models to “think out loud” \emph{reduces} accuracy on tasks that rely on quick visual yes/no or single-label judgments, such as scene transition detection. The additional tokens invite over-rationalization, hallucinated details, and context-length dilution—turning what should be a fast lookup into an opportunity to override the correct first impression. In contrast, CoT \emph{improves} accuracy on more cognitively demanding tasks, like counting object movements or actions. As shown in Figure~\ref{fig:cot_vs_only_answer_pie}, even with the much larger proprietary model GPT-4o—a strong model with advanced reasoning capabilities—a significant portion of questions are better answered directly than with CoT. 

\begin{figure}[t!]
    \centering
    \includegraphics[width=0.95\textwidth]{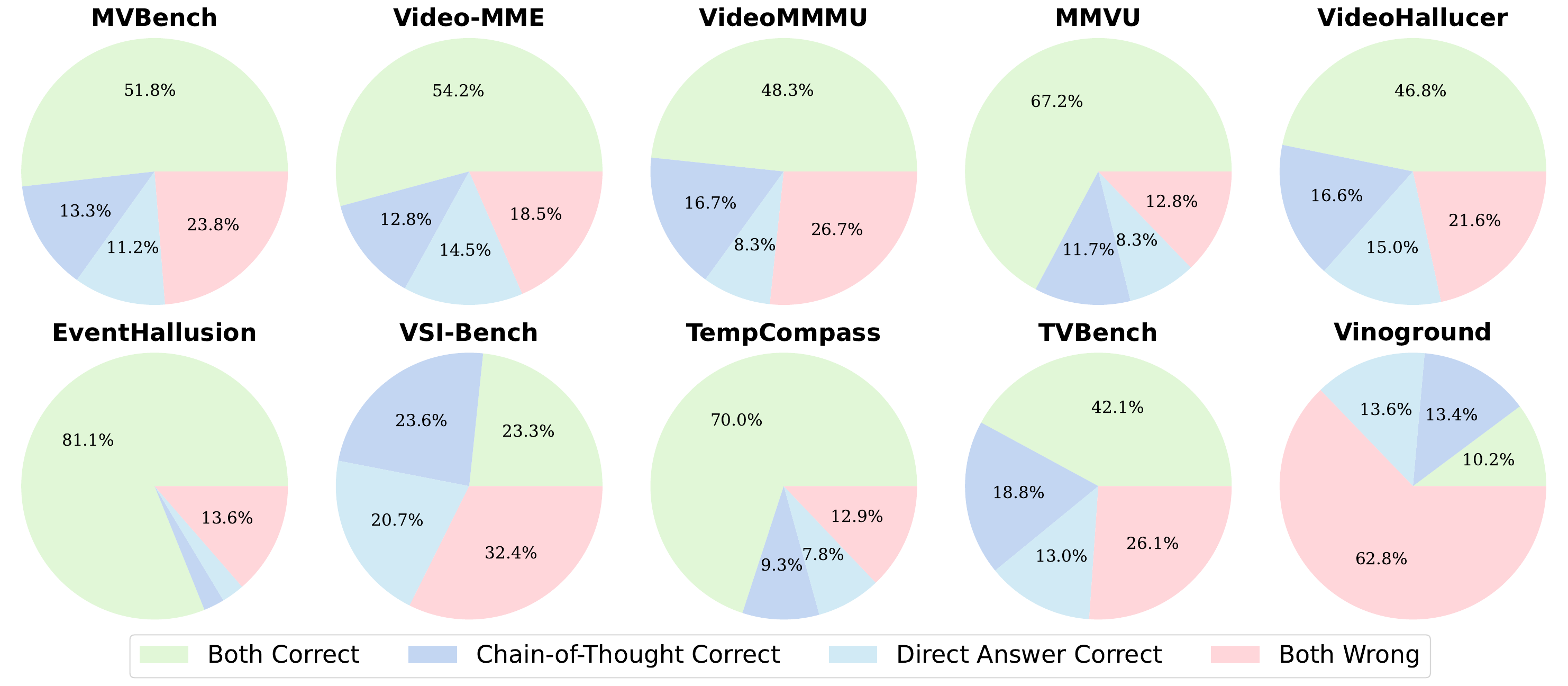}
    \caption{Even with GPT-4o (a strong reasoning model), a considerable portion of questions (light blue areas) are better answered directly than with CoT reasoning, implying significant room for improvement in CoT reasoning. For VSI-Bench and MMVU, results are based on MCQ subset.
}
    \label{fig:cot_vs_only_answer_pie}
\end{figure}

In summary, while reasoning-driven generation is promising for complex tasks requiring explicit decomposition, it can hinder performance on simpler perceptual tasks by introducing unnecessary reasoning steps, leading to errors or hallucinations.

\subsection{Visual Thinking Drift: When Reasoning Ignores the Video}
\label{sec:visual_thinking_drift}
Reasoning implicitly unfolds in two stages: first, identifying the relevant rules and facts needed to reach a conclusion, and second, applying them effectively to arrive at that conclusion~\cite{levesque1989knowledge}. Simply encoding knowledge is not enough—robust reasoning under uncertainty is essential. However, the verbose nature of CoT traces introduces stochasticity into the reasoning process. Drawing inspiration from self-consistency~\cite{wang2022self}, we found that majority voting over 20 responses generated with CoT prompt significantly improves accuracy (see supplementary material for details). Yet, this improved stability comes at the cost of considerable computational overhead. 

To investigate the source of this instability, we analyzed multiple erroneous chains of thought. Paradoxically, we found that many flawed thinking traces in video analysis are logically flawless. The culprit? A phenomenon we call “\textbf{Visual Thinking Drift}”, illustrated in Figure~\ref{fig:visual_thinking_drift}. The model's reasoning is sound, but it is unmoored from the video's true content—building its logic on hallucinated visual details or isolated temporal fragments, which inevitably steer the inference off track.

To better understand the Visual Thinking Drift phenomenon, we adopt a Bayesian lens, which helps  disentangle why CoT can damage a video LLM’s answer even when the direct answer alone is correct. Consider a video LLM that, given a question $q$ and video features $\mathbf v$,
generates a chain of reasoning tokens $c_{1:T}$ followed by a final answer $a$.
Its implicit generative story is
\[
  p(c_{1:T},a \mid q,\mathbf v)
  \;=\;
  p\!\bigl(a \mid c_{1:T},q,\mathbf v\bigr)
  \prod_{t=1}^{T} p\!\bigl(c_t \mid c_{<t},q,\mathbf v\bigr).
\]
Because the chain tokens are never supervised, each inference step samples a
high‑variance \emph{latent} state.  Early in the generation process, visual
evidence does influence the softmax
\[
  p\!\bigl(c_t \mid c_{<t},q,\mathbf v\bigr)\;\propto\;
  \exp\bigl(
    \underbrace{\mathbf h_{c_{<t}}^{\!\top}W_{\text{lang}}}_{\text{language prior}}
    +
    \underbrace{\mathbf h_{\mathbf v}^{\!\top}W_{\text{vis}}}_{\text{visual likelihood}}
  \bigr),
\]
yet in practice $\lVert W_{\text{lang}}\rVert \!\gg\! \lVert W_{\text{vis}}\rVert$.
As $t$ grows, self‑attention focuses ever more on the already‑generated tokens,
so the visual likelihood is \emph{diluted}.  A linguistically plausible but
ungrounded narrative emerges, and a single hallucinated detail can dominate all
subsequent probabilities.

If each individual step is correct with probability $p_s = 1-\varepsilon$, the
probability that an entire chain of length $T$ is error‑free is
$(1-\varepsilon)^{T}\!\approx 1-T\varepsilon$ for small~$\varepsilon$; thus the
failure rate grows \emph{linearly} with chain length.  Once an early token
commits to a nonexistent visual fact (e.g.\ “the man holds a red ball”), all
later tokens and the answer are conditioned on this fiction.  Because
autoregressive decoding has no backward message‑passing to re‑verify the video,
the posterior mass collapses around the hallucination and recovery becomes
nearly impossible.

Crucially, the model’s high‑probability \emph{logical scaffolds}—“if–then”
structures, counting loops, temporal ordering—stay intact, while the low‑entropy visual details are weakly weighted.  The result is a chain of thought that \emph{looks} logically sound but is built on visually hallucinated content: the essence of \emph{visual thinking drift}.

We show two concept examples of visual thinking drift made by CoT prompting in Figure~\ref{fig:visual_thinking_drift}, where CoT brings hallucinations or mismatches with visual evidence. When inconsistencies arise, MLLMs faithfully trust textual data over visual data, leading to wrong reasoning paths.

\section{Visual Evidence Reward (VER) for MLLM Video Reasoning}
\begin{figure}[t]
    \centering
    \includegraphics[width=1\textwidth]{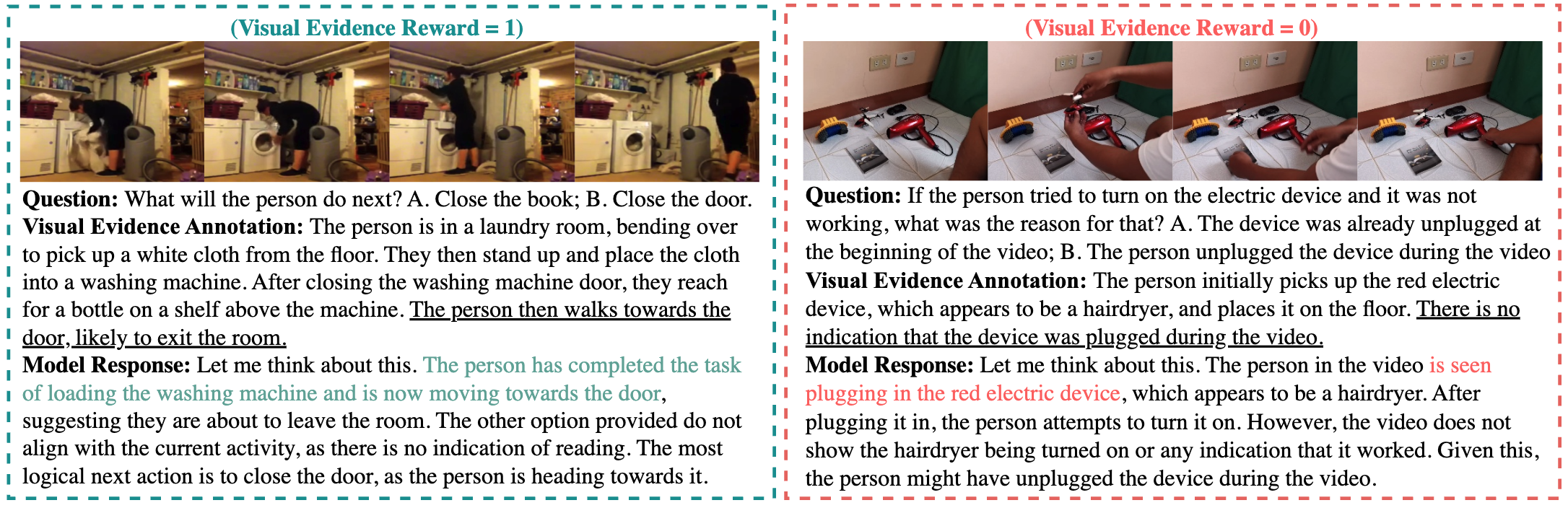}
    \caption{Visualization of visual facts generated from the training data. Chain-of-thought responses that actively reference visual evidence are rewarded, while those that do not receive zero reward.}\vspace{-5pt}
    \label{fig:visual_fact_generation}
\end{figure}
\noindent

Continuing this Bayesian perspective, a key insight into the visual thinking drift dilemma is that CoT tokens are never explicitly supervised during training. Without clear guidance, reasoning can easily drift away from the visual evidence it is meant to be grounded in. To tackle this, we enhance the lightweight rule-based reinforcement learning algorithm, Group Relative Policy Optimization (GRPO)~\cite{shao2024deepseekmath, guo2025deepseek}, by introducing a novel reward mechanism: Visual Evidence Reward (VER). VER directly supervises the model’s reasoning process by rewarding it when its chain-of-thought includes accurate and relevant visual details—effectively anchoring abstract reasoning in concrete visual facts.

For each question~$q$, we have a policy model $\pi_\theta$ to generate a group of responses~$\{o_i\}_{i=1}^{G}$, where $G$ is the group size. A large language model (LLM)-based judge evaluates each response~$o_i$ for its reference to the visual evidence~$\mathbf{v}$, assigning a binary score $e_i \in \{0,1\}$, where $1$ indicates a proper reference. We define the evidence reward coefficient as $r_e = \alpha$ if $e_i = 1$, and $r_e = 0$ otherwise, where $\alpha$ is a tunable reward weight. Note that using an auxiliary LLM to generate similarity scores for reward calculation is a common practice in recent work~\cite{zheng2023judging,gu2024survey,xie2023text2reward} in the language domain.

The evidence-augmented reward is computed as $r_i^{\text{evid}} = r_i + r_e$ if both $o_i$ is correct and $e_i = 1$; otherwise, $r_i^{\text{evid}} = r_i$. For each question, we compute the group reward $\mathbf{r} = \{r_i^{\text{evid}}\}_i^G$, and use it to normalize the rewards: $A_i = \frac{r_i^{\text{evid}} - \mathrm{mean}(\mathbf{r})}{\mathrm{std}(\mathbf{r})}$.

With this, the policy is optimized via the clipped GRPO objective:
\begin{align*}
\mathcal{J}_{\text{evid-GRPO}}(\theta) 
= &\mathbb{E}_{q,\, \{o_i\}_{i=1}^{G} \sim \pi_{\theta_{\text{old}}}(O \mid q)} \Bigg[ 
\frac{1}{G} \sum_{i=1}^G \Bigg(
\min\Bigg(
\frac{\pi_\theta(o_i \mid q)}{\pi_{\theta_{\text{old}}}(o_i \mid q)} A_i,\,
\\
&\text{clip}\left(
  \frac{\pi_\theta(o_i \mid q)}{\pi_{\theta_{\text{old}}}(o_i \mid q)},
  1 - \epsilon,\,
  1 + \epsilon
\right) A_i
\Bigg)
- \beta\, \mathbb{D}_{\mathrm{KL}}\big(\pi_\theta \,\|\, \pi_{\mathrm{ref}}\big)
\Bigg) 
\Bigg]
\end{align*}

where $\epsilon$ and $\beta$ are hyperparameters for clipping and KL regularization, respectively. $\pi_\theta$ denotes the current policy, $\pi_{\theta_\text{old}}$ is the prior policy used for importance sampling, and $\pi_{\theta_\text{ref}}$ is the reference model set to the initial checkpoint for regularization.

By incorporating an evidence-based reward signal, \methodname~explicitly encourages models to ground their reasoning in visual context, leading to more contextually relevant responses.

\custompar{Visual Evidence Generation}
What exactly qualifies as visual evidence? A straightforward approach might be to use general video captions~\cite{abdar2024review}—but this quickly runs into the issue of granularity. Captions often miss the specific visual cues needed to answer particular questions. To resolve this, we define visual evidence in a question-dependent manner: it consists of the visual details necessary to answer a given question, which can vary significantly across tasks.

To generate such evidence, we leverage a strong MLLM, Qwen2.5-VL-72B~\cite{Qwen2.5-VL}, prompting it with both the video and the question. The model is asked to produce not only an answer but also a list of visual observations that support that answer. This way, the reasoning process remains tightly grounded in relevant visual input. Full prompt details are provided in the supplementary materials. Qualitative examples of the generated visual evidence used for GRPO training are shown in Figure~\ref{fig:visual_fact_generation}. 

The external VLM is only used to generate training data in the form of question-specific visual evidence. Our policy model (\modelname), trained on this evidence, performs inference independently and does not rely on the external VLM at test time.
We recognize that VLM-generated outputs may include speculative or hallucinated content. To mitigate this, we filter and structure the visual evidence via carefully designed prompts (see supplementary materials), ensuring that the extracted visual details are question-relevant and textually explicit. Empirically, this approach results in higher alignment between reasoning chains and observable video content, as demonstrated in Figure~\ref{fig:qualitative}.
\section{Experiments}
\label{sec:exp}
\custompar{Training Strategy}
Our model is a post-trained Qwen2.5-VL-7B \cite{bai2025qwen2}, employing a two-stage pipeline that combines supervised fine-tuning (SFT) and reinforcement learning (RL). 
The process begins with SFT, where we train the model on Video-R1-COT-165k dataset~\cite{feng2025videor1}, a dataset providing chain-of-thought (CoT) annotations, helping bootstrap the model’s reasoning abilities during the cold-start phase. Following this, the model undergoes reinforcement learning using GRPO with our \methodname~(VER). This phase uses a dataset mixture comprising Reversed-in-Time~\cite{Du2024RTime} and Video-R1-260k~\cite{feng2025videor1} samples. The RL stage is designed to move the model beyond the constraints of supervised learning, allowing it to develop more robust and adaptable reasoning patterns through exploration and self-guided optimization. We call our final post-trained model \modelname.

\custompar{Benchmarks}
We extensively evaluate our model across a broad spectrum of 10 public video understanding benchmarks, covering a wide range of reasoning skills. These include comprehensive, all-around benchmarks such as MVBench~\cite{li2024mvbench} and Video-MME~\cite{fu2024videomme}; temporal reasoning benchmarks like TVBench~\cite{cores2024tvbench}, Vinoground~\cite{zhang2024vinoground}, and TempCompass~\cite{liu2024tempcompass}; spatial reasoning benchmarks such as VSI-Bench~\cite{yang2024vsibench}; and knowledge-intensive datasets including Video-MMMU~\cite{hu2025videommmu} and MMVU~\cite{zhao2025mmvu}. We also assess robustness to hallucination using dedicated benchmarks such as EventHallusion~\cite{eventhallusion} and VideoHallucer~\cite{videohallucer}. Most benchmarks consist of multiple-choice questions (MCQs), with the exception of VideoMMMU and VSI-Bench, which include questions requiring numerical answers. For MMVU, we follow the protocol from~\cite{feng2025videor1} and use its MCQ subset to ensure robust and consistent evaluation. Subtitles are excluded in the Video-MME evaluation setting.

\begin{figure}[t]
    \centering
    \includegraphics[width=1\textwidth]{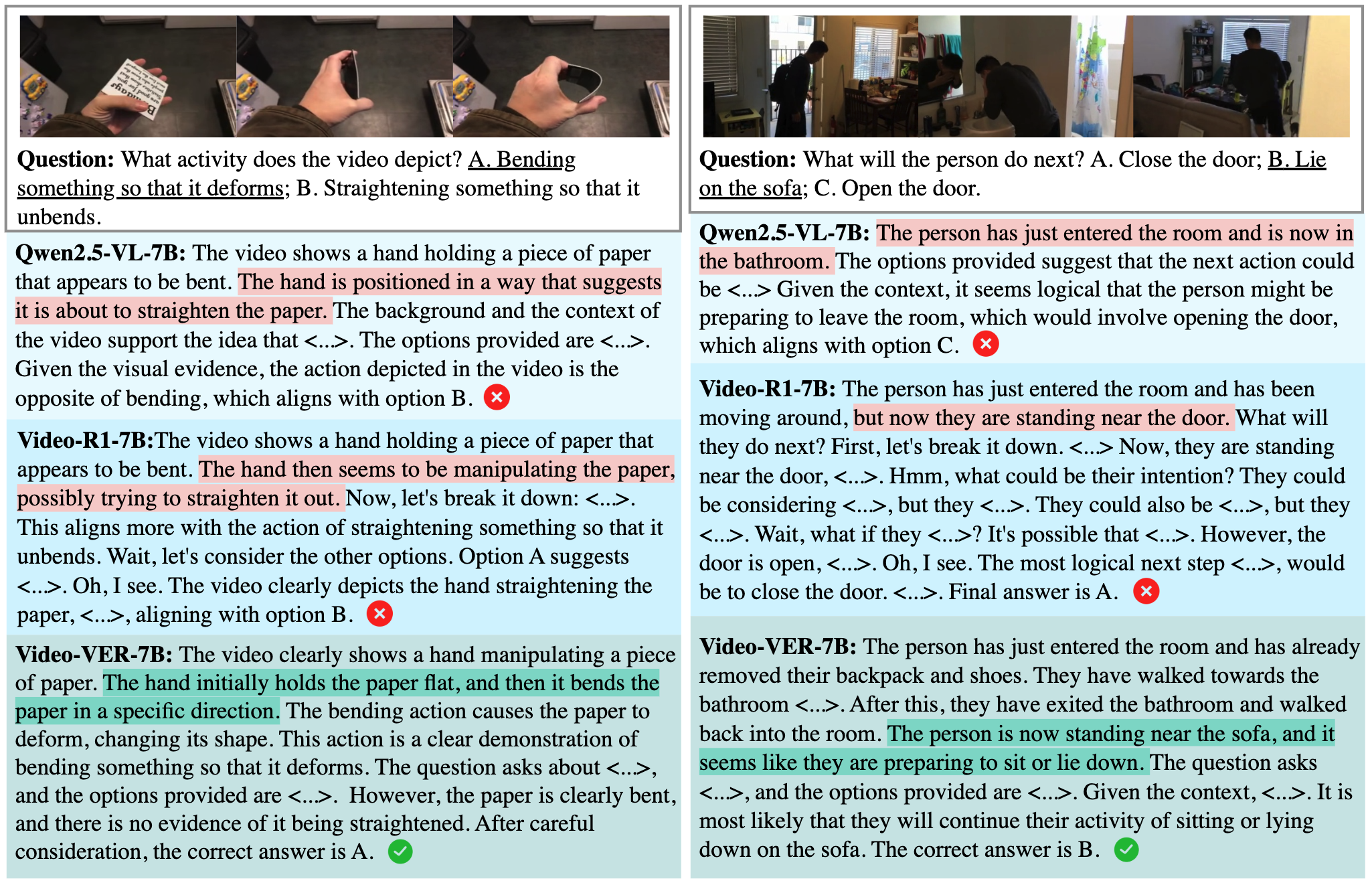}
    \caption{Comparison of reasoning traces from baselines and  our \modelname ~model. Notice how baseline models often include speculative or hallucinated details not grounded in the video, whereas \modelname~maintains alignment between intermediate reasoning steps and observable evidence.}
    \label{fig:qualitative}
\end{figure}
\noindent

\begin{table}[t] 
\centering
\caption{Comparison of \modelname~with baselines across 10 video benchmarks. Our model consistently ranks \best{first} or \sbest{second} overall, demonstrating the effectiveness of evidence-grounded chain-of-thought (CoT) reasoning. Notably, across most base models (e.g., Qwen2.5-VL), CoT prompting often leads to lower accuracy compared to direct answering (DA), highlighting the risk of ungrounded reasoning. In contrast, \modelname~maintains or improves performance with CoT by explicitly grounding reasoning in question-relevant visual evidence. To emphasize this, we annotate the accuracy margins (↑) between \modelname~and its base model Qwen2.5-VL-7B---both with CoT---in \textcolor{ForestGreen}{small offset font}, drawing attention to the consistent gains enabled by our visual grounding reward.
}
\resizebox{\textwidth}{!}{%
\begin{tabular}{lllllllll}
\thickhline
\multirow{2}{*}{Model}  & \multirow{2}{*}{Size}  & \multirow{2}{*}{Prompt}  &\multirow{2}{*}{\textbf{MVBench}} & \multirow{2}{*}{\textbf{Video-MME}} & \multirow{2}{*}{\textbf{VideoMMMU}} & \multirow{2}{*}{\textbf{MMVU}}  & \multirow{2}{*}{\textbf{VideoHallucer}} & \multirow{2}{*}{\textbf{EventHallusion}} \\
 & & & &  & &  & \\
 \hline
\rowcolor{LighterGray}
GPT-4o~\cite{achiam2023gpt} & - & DA & 62.9 & 68.7 & 56.7 & 75.5 & 61.8 & 83.9 \\
\rowcolor{LighterGray}
GPT-4o~\cite{achiam2023gpt} & - & COT & 65.1 & 67.0 & 65.0 & 78.9 & 63.4 & 83.6 \\

LongVA~\cite{zhang2024longva} & 7B & DA & -& 52.6 & 23.9 & - & - & - \\
Video-UTR~\cite{yu2025unhackable} & 7B & DA & 58.8 & 52.6 & - & - & - & - \\
LLaVA-OneVision~\cite{li2024llava} & 7B  & DA & 57.1 & 57.7 & 33.8 & 49.2 & 34.7 & 61.1 \\
Kangeroo~\cite{liu2024kangaroo} & 8B  & DA & 61.1 & 56.0 & - & - & - & - \\
TinyLLaVA-Video-R1~\cite{zhang2025tinyllava} & 3B & COT & 49.5 & 46.6 & - & 46.9 & - & -\\
Video-R1~\cite{feng2025videor1} & 7B & COT & \sbest{63.9} & \best{59.3} & \sbest{52.3} & 63.8 & 44.2 & 51.4 \\
\hline
Qwen2.5-VL~\cite{Qwen2.5-VL} & 3B & DA & 62.8 & 55.9 & 41.7 & 59.5 & 44.8 & 54.9 \\
Qwen2.5-VL~\cite{Qwen2.5-VL} & 3B & COT & 57.4 & 53.2 & 33.8 & 53.6 & 40.4 & 49.9 \\
Qwen2.5-VL~\cite{Qwen2.5-VL} & 7B & DA & 63.6 & \sbest{59.2} & 47.3 & \sbest{64.2} & \sbest{51.8} & 64.5 \\
Qwen2.5-VL~\cite{Qwen2.5-VL} & 7B & COT & 59.8 & 54.7 & 47.8 & 60.5 & 44.1 & \sbest{67.3} \\
\hline
\rowcolor{LightPurple}
\modelname~(Ours) & 7B & COT 
& \best{64.1}~{\textcolor{ForestGreen}{\footnotesize (+4.3)}} 
& \best{59.3}~{\textcolor{ForestGreen}{\footnotesize (+4.6)}} 
& \best{52.7}~{\textcolor{ForestGreen}{\footnotesize (+4.9)}} 
& \best{65.1}~{\textcolor{ForestGreen}{\footnotesize (+4.6)}} 
& \best{53.1}~{\textcolor{ForestGreen}{\footnotesize (+9.0)}} 
& \best{70.0}~{\textcolor{ForestGreen}{\footnotesize (+2.7)}} \\

\thickhline
\multirow{2}{*}{Model}  & \multirow{2}{*}{Size}  & \multirow{2}{*}{Prompt} & \multirow{2}{*}{\textbf{VSI-Bench}} & \multirow{2}{*}{\textbf{TempCompass}} & \multirow{2}{*}{\textbf{TVBench}} & \multicolumn{3}{c}{\textbf{Vinoground}}\\
\cline{7-9}
\textbf{} & &  & & & & \textbf{Text} & \textbf{Video} &  \textbf{Group}\\
\hline
\rowcolor{LighterGray}
GPT-4o~\cite{achiam2023gpt} & - & DA & 27.8 & 77.8 & 55.1 & 57.6 & 34.4 & 23.8 \\
\rowcolor{LighterGray}
GPT-4o~\cite{achiam2023gpt} & - & COT & 45.3 & 79.3 & 60.9 & 62.2 & 38.0 & 23.6 \\
LongVA~\cite{zhang2024longva} & 7B & DA & - & 56.9 & - & - & - & - \\
Video-UTR~\cite{yu2025unhackable} & 7B & DA & - & 59.7 & - & - & - & - \\
LLaVA-OneVision~\cite{li2024llava} & 7B  & DA & 32.9 & 67.8 & 47.2 & 42.0 & 28.4 & 12.8 \\
Kangeroo~\cite{liu2024kangaroo} & 8B  & DA & - & 62.5 & - & - & - & - \\
Video-R1~\cite{feng2025videor1} & 7B & COT & \best{35.8} & 73.2 & \sbest{52.4} & 34.6 & 24.8 & 9.4 \\
\hline
Qwen2.5-VL~\cite{Qwen2.5-VL} & 3B & DA & 30.1 & 67.3 & 45.2 & 30.2 & 21.2 & 6.0 \\
Qwen2.5-VL~\cite{Qwen2.5-VL} & 3B & COT & 25.4 & 59.8 & 47.2 & 30.8 & 22.6 & 5.8 \\
Qwen2.5-VL~\cite{Qwen2.5-VL} & 7B & DA & 32.3 & \sbest{73.7} & 52.2 & \sbest{42.2} & \best{29.2} & \best{14.8} \\
Qwen2.5-VL~\cite{Qwen2.5-VL} & 7B & COT & 31.4 & 71.3 & 49.9 & 40.6 & 28.0 & 12.8 \\
\hline
\rowcolor{LightPurple}
\modelname~(Ours) & 7B & COT 
& \sbest{34.6}~{\textcolor{ForestGreen}{\footnotesize (+3.2)}} 
& \best{74.0}~{\textcolor{ForestGreen}{\footnotesize (+2.7)}} 
& \best{52.8}~{\textcolor{ForestGreen}{\footnotesize (+2.9)}} 
& \best{42.8}~{\textcolor{ForestGreen}{\footnotesize (+2.2)}} 
& \sbest{28.2}~{\textcolor{ForestGreen}{\footnotesize (+0.2)}} 
& \sbest{14.4}~{\textcolor{ForestGreen}{\footnotesize (+1.6)}} \\
\thickhline
\end{tabular}
}
\label{tab:main_results}
\end{table}

\custompar{Implementation Details}
During training, the maximum number of video frames is set as 16, and increased to 32 at inference time for both our model and all baselines, unless otherwise specified. For GRPO training, we incorporate four reward components: an accuracy reward, our visual evidence reward (with weight $\alpha = 0.3$), a format reward to encourage consistent answer structure, and a length reward to promote moderately long, informative responses. We train our model with 8 NVIDIA H200 GPUs.  GRPO group size $G$ is set as 8. The number of RL iterations is set to 2,000. Further implementation details can be found in the supplementary material.

\custompar{Baselines}
We compare our model against both proprietary and open-source video LLMs. For the proprietary model, we evaluate GPT-4o~\cite{hurst2024gpt}. For open-source baselines, we evaluate against LongVA~\cite{zhang2024longva}, Video-UTR~\cite{yu2025unhackable}, LLaVA-OneVision~\cite{li2024llava}, Kangaroo~\cite{liu2024kangaroo}, and Qwen2.5-VL~\cite{Qwen2.5-VL}. Additionally, to assess reasoning capabilities, we include two recently released video reasoning models: TinyLLaVA-Video-R1~\cite{zhang2025tinyllava} and Video-R1~\cite{feng2025videor1}, both of which are explicitly designed for text-based multi-step video reasoning tasks. Together, these baselines span the full spectrum of contemporary video-language systems---from large, commercially deployed models to lean, community-driven releases---ensuring that our evaluation is both representative of the strongest available competitors and informative for researchers and practitioners who rely on open tools.

\subsection{Evidence-Grounded Chains Lead to Better Video Understanding}
Table~\ref{tab:main_results} demonstrates that \modelname~consistently surpasses existing open-source video MLLMs across a comprehensive suite of benchmarks, ranking first in 9 out of 10 evaluations, except for the VSI-Bench where Video-R1~\cite{feng2025videor1} achieves the strongest performance.  \modelname~shows superior results in temporally nuanced data like TempCompass (74.0\%) and TVBench (52.8\%), underscoring its ability to interpret sequential information and dynamic visual content with precision. These results validate the strength of our proposed method and its generalization across diverse video-language tasks, all while utilizing a 7B parameter model with an innovative RL training strategy. 

Figure~\ref{fig:qualitative} presents qualitative examples of the thinking chains generated by our model and baseline methods. Some raw text has been omitted for brevity. The results illustrate that the reasoning of baseline models is often distracted by speculative or hallucinated details, which are not grounded in the actual actions or the full temporal context of the video. In contrast, \modelname~better maintains alignment between intermediate reasoning steps and observable evidence, leading to the correct answer. We discuss common \textbf{failure cases} and \textbf{limitations} of our~\modelname~in the appendix.

\subsection{Ablation Study}
\begin{table}[h!]
\centering
\caption{Ablation on types of visual evidence, showing that question-dependent visual evidence (QD-VE) is generally preferred over general video captions (VC) for visual evidence generation.}
\resizebox{\textwidth}{!}{%
\begin{tabular}{llllllllllllll}
\thickhline
\multirow{2}{*}{Model} & \multirow{2}{*}{\#Type} & \multirow{2}{*}{\textbf{MVBench}} & \multirow{2}{*}{\textbf{Video-MME}} & \multirow{2}{*}{\textbf{VideoMMMU}} & \multirow{2}{*}{\textbf{MMVU}} & \multirow{2}{*}{\textbf{VideoHal.}} & \multirow{2}{*}{\textbf{EventHal.}} & \multirow{2}{*}{\textbf{VSI-Bench}} & \multirow{2}{*}{\textbf{TempC.}} & \multirow{2}{*}{\textbf{TVBench}} & \multirow{2}{*}{\textbf{Vinog.}} \\
& & & & & & & & & & & \\ 
\hline
\rowcolor{LightPurple}
\modelname & QD-VE & \best{64.1} & \best{59.3} & \best{52.7} & \best{65.1} & \best{53.1} & 70.0 & \best{34.6} & \best{74.0} & \best{52.8} & \best{14.4} \\
\modelname & VC & 63.9 & 58.7 & 52.2 & 64.8 & 52.5 & \best{70.3} & 34.4 & 73.6 & 52.4 & 13.6 \\
\thickhline
\end{tabular}
}
\label{tab:ve_type} 
\end{table}

\begin{table}[h!]
\centering
\caption{Ablation study on the scalability of \modelname~across varying temporal context lengths.}
\resizebox{\textwidth}{!}{%
\begin{tabular}{llcccccccccc}
\thickhline
\multirow{2}{*}{Model} & \multirow{2}{*}{\#Frames} & \multirow{2}{*}{\textbf{MVBench}} & \multirow{2}{*}{\textbf{Video-MME}} & \multirow{2}{*}{\textbf{VideoMMMU}} & \multirow{2}{*}{\textbf{MMVU}} & \multirow{2}{*}{\textbf{VideoHal.}} & \multirow{2}{*}{\textbf{EventHal.}} & \multirow{2}{*}{\textbf{VSI-Bench}} & \multirow{2}{*}{\textbf{TempC.}} & \multirow{2}{*}{\textbf{TVBench}} & \multirow{2}{*}{\textbf{Vinog.}} \\
& & & & & & & & & & & \\ 
\hline
\rowcolor{LightPurple}
\modelname & 32 & \best{64.1} & \best{59.3} & \best{52.7} & \best{65.1} & \best{53.1} & 70.0 & 34.6 & \best{74.0} & \best{52.8} & \best{14.4} \\
\modelname & 16 & 63.2 & 56.0 & 50.0 & 64.8 & 51.4 & 69.8 & \best{35.2} & 72.8 & 51.0 & 10.6 \\
\modelname & 8  & 60.5 & 53.3 & 45.4 & 63.2 & 50.4 & \best{70.5} & 32.6 & 69.6 & 48.3 & 11.0 \\
\thickhline
\end{tabular}
}
\label{tab:frames} 
\end{table}
\custompar{Types of Visual Evidence}
As shown in Table~\ref{tab:ve_type}, we experiment with two approaches to generating visual evidence to train our \modelname~model: question-dependent visual evidence (QD-VE), obtained by prompting a MLLM with both the question and the video, and a generic video caption (VC), generated by asking the MLLM to produce a detailed caption of the video alone. The Group score is reported for the Vinoground benchmark. Our ablation shows that QD-VE outperforms VC on 9 out of 10 benchmarks, and falls behind only on EventHallusion. These results highlight the effectiveness of aligning visual evidence with the specific question, confirming that question-tailored context is more beneficial than a general-purpose description. 
\custompar{Scalability with Frames}
Table \ref{tab:frames} presents an ablation study assessing the scalability of \modelname~with varying numbers of input frames. The Group score is reported for the Vinoground benchmark. The results reveal a consistent trend: performance improves as more frames are provided, with the 32-frame setting achieving the best results on 8 out of 10 benchmarks. This demonstrates that our method effectively leverages more frames with longer temporal context, confirming its ability to scale and generalize well across different video lengths.

\section{Conclusions and Future Work}
Bridging the gap between the linear, symbolic structure of chains-of-thought and the inherently fuzzy, non-linear, and temporally distributed nature of video remains a central challenge. This paper has highlighted a critical failure mode in this pursuit: “\emph{Visual Thinking Drift}”, where reasoning processes, despite appearing coherent, become unmoored from actual video content. We then introduced Visual Evidence Reward (VER), a novel reinforcement learning reward mechanism specifically designed to counteract this drift. Our VER excels in closed-ended tasks like multiple-choice question answering where rule-based reward computation is effective. Extending this framework to open-ended tasks, such as free-form QA, and thereby ensuring that verbosity is replaced by genuinely grounded video intelligence, presents an exciting and crucial avenue for future work.}

\newpage
\clearpage
{
    \small
    \bibliographystyle{plain}
    \bibliography{main}
}

\newpage
\appendix
\tableofcontents

\section{Self-consistency Decoding for Video Reasoning}
Motivated by the decoding strategy self-consistency proposed in~\cite{wang2022self}, which samples a diverse set of reasoning paths instead of relying solely on greedy decoding, and then selects the most consistent answer by marginalizing over the sampled paths, we explore its implications for video reasoning tasks. The intuition behind self-consistency is that complex reasoning problems often allow for multiple valid reasoning trajectories that converge on a unique correct answer.

However, our focus is on video reasoning tasks, which generally exhibit lower reasoning complexity than language-only tasks such as mathematical problem solving or code generation. Moreover, the diversity of reasoning paths in video-based tasks tends to be more constrained due to the fixed visual context and limited temporal narrative. For example, while a math problem might allow several logical formulations or decompositions, a video clip typically presents a specific sequence of events that restricts interpretive variation.

\begin{figure}[ht]
    \centering
    \includegraphics[width=0.99\textwidth]{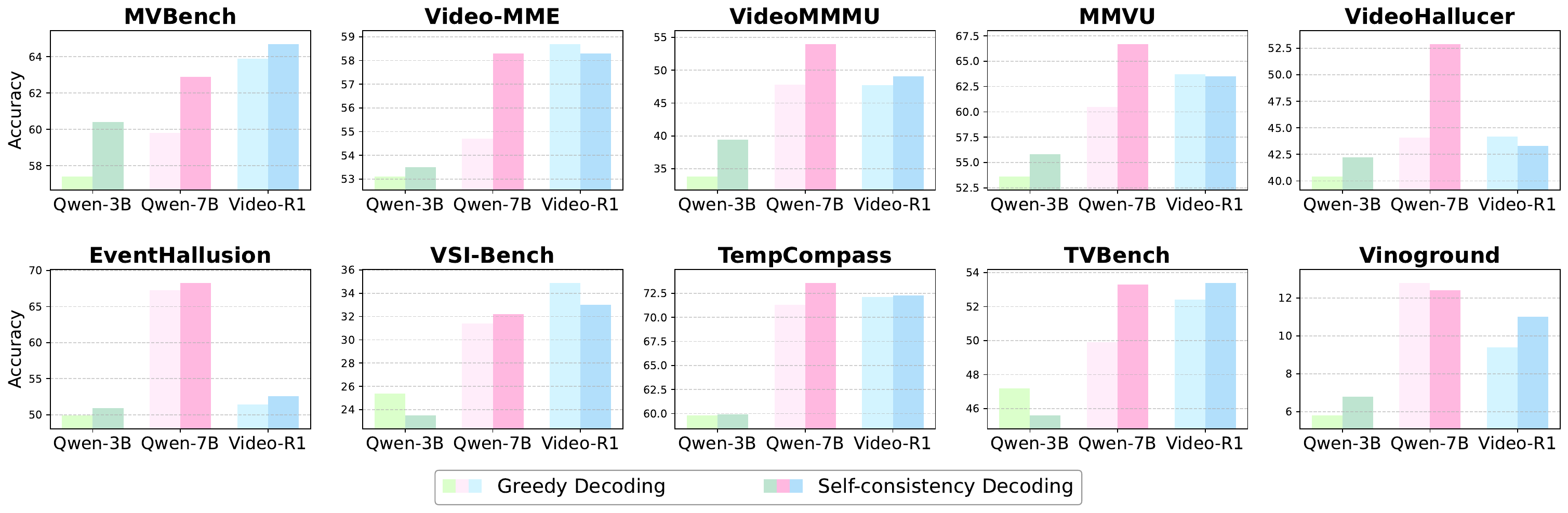}
    \caption{Sampling multiple independent CoT responses per question and aggregating them via majority voting yields a clear accuracy improvement---indicating that reasoning traces are often stochastic rather than dependable. Each chart is a different popular video benchmark.}
    \label{fig:bar_major_vote}
\end{figure}

Despite this, we observe that simple majority voting over 20 responses generated using Chain-of-Thought (CoT) prompting  (with the same model)
significantly boosts accuracy across all models in most scenarios. See Figure~\ref{fig:bar_major_vote}. In particular, this indicates that the reasoning traces for video tasks are often unstable, and greedy decoding is more prone to getting trapped by the \emph{visual thinking drift}—the phenomenon discussed in Section~\ref{sec:visual_thinking_drift}, where subtle ambiguities or misinterpretations in the visual context derail the reasoning process. By aggregating multiple responses, self-consistency helps to smooth out these drifts and converge on more robust answers.

\section{Implementation Details}

\custompar{Training and Testing Configurations}
For evaluation, we set the temperature to 0.01 for all baseline models as well as our model. During training, the maximum video token budget is set to 128 × 28 × 28 pixels.
During testing, this is increased to 256 × 28 × 28 pixels.
The sampling rate is set to 2.0 FPS across all benchmarks, except for Vinoground, which uses 4 FPS.
To manage API costs, the maximum number of frames used by GPT-4o is limited to 16 frames per video, except for TempCompass, where only 8 frames are used.

The results for VSI-Bench reported in Table~\ref{tab:main_results} represent the average accuracy across both multiple-choice and regression-based tasks.
For the baseline model Video-R1, we adopt the reported results of MVBench, Video-MME, VideoMMMU, MMVU, VSI-Bench, and TempCompass under the 32-frame setting as specified in their respective papers. For the remaining four benchmarks—VideoHallucer, EventHallusion, TVBench, and Vinoground—we conduct our own evaluations due to the absence of publicly available results. In Figure~\ref{fig:bar_cot_harm} and Figure~\ref{fig:bar_major_vote}, to ensure a fair and consistent comparison across all prompting and decoding strategies, we re-evaluate Video-R1 on all benchmarks in our experimental setting, including those with reported numbers, wherever applicable.

\custompar{Length Reward}
To encourage deeper reasoning without excessive verbosity, we apply a length reward targeting response lengths in the range of 320 to 512 tokens.

\custompar{LLM Judge}
We utilize Llama-3.1-70B-Instruct~\cite{grattafiori2024llama} as our LLM-based judge. It is prompted to produce a binary label—1 for successfully referencing the visual evidence, and 0 for failing to do so.

\begin{figure}[t!]
\centering
\includegraphics[width=0.99\textwidth]{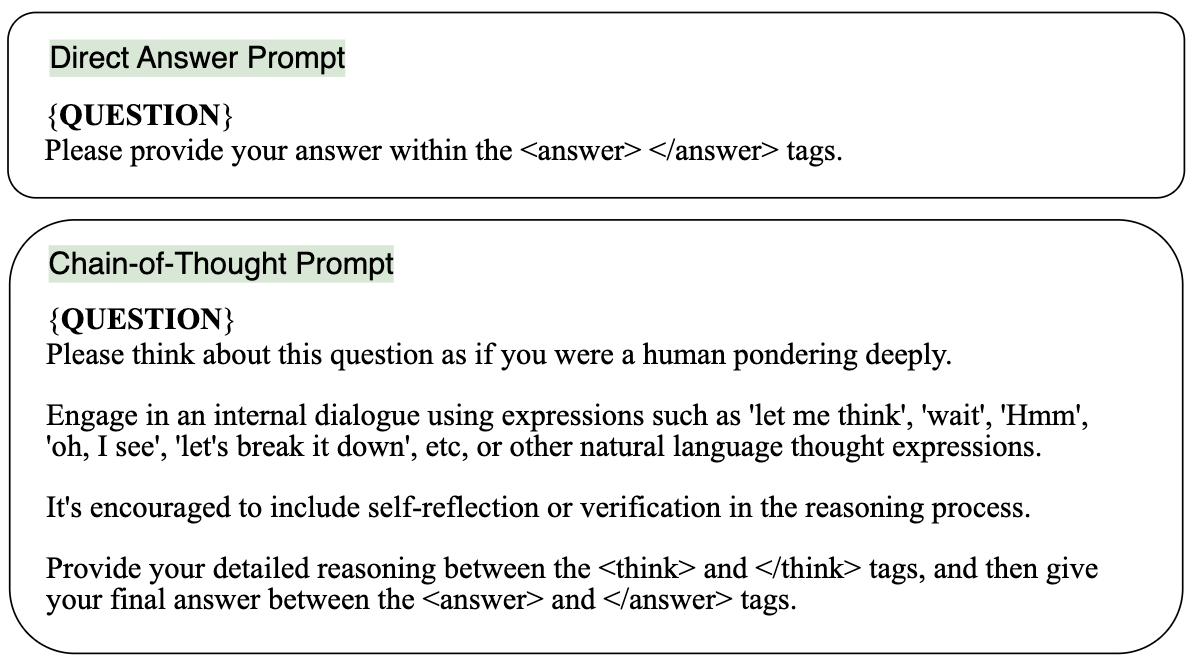}
\caption{Prompts used for direct response generation and reasoning-driven generation (chain-of-thought). The CoT prompt is borrowed from Video-R1~\cite{feng2025videor1}.}
\label{fig:prompt_direct_vs_cot}
\end{figure}

\begin{figure}[t!]
\centering
\includegraphics[width=0.99\textwidth]{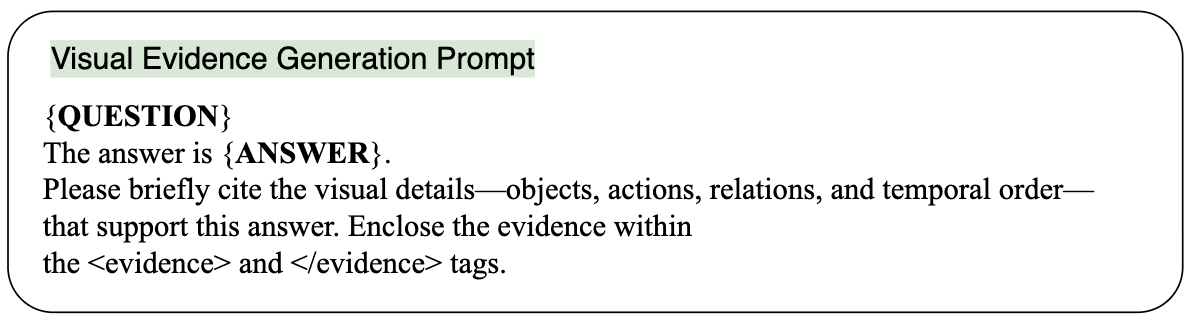}
\caption{Inverted Prompting used for generating visual evidence annotations.}
\label{fig:prompt_ver}
\end{figure}

\custompar{Prompt Details}
All prompts used in our training and evaluation are illustrated in Figures~\ref{fig:prompt_direct_vs_cot} and~\ref{fig:prompt_ver}.

\FloatBarrier
\section{LLM-Based Judge for Visual Evidence Grounding}
Evaluating whether a model’s reasoning references the correct visual evidence is inherently a semantic task that goes beyond exact string or token overlap. It requires assessing whether the generated rationale mentions visual facts that are relevant, specific, and consistent with what is shown in the video. To enable this, we adopt an auxiliary LLM-based judge to compute a binary reward—assigning 1 if the rationale includes grounded, question-relevant visual details, and 0 otherwise.

\begin{figure}[t]
\centering
\includegraphics[width=0.99\textwidth]{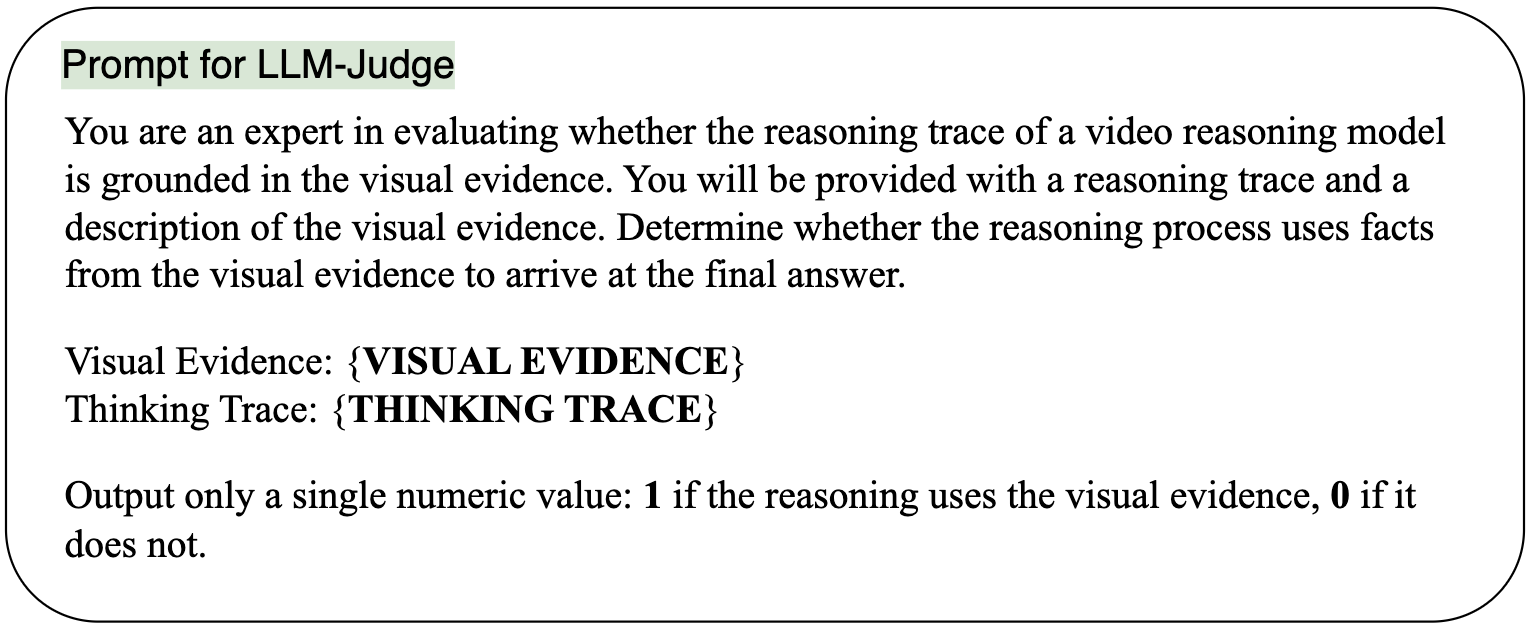}
\caption{The prompt used for LLM Judge, producing reward value 0 or 1.}
\label{fig:prompt_llm_judge}
\end{figure}

\begin{figure}[t!]
\centering
\includegraphics[width=0.99\textwidth]{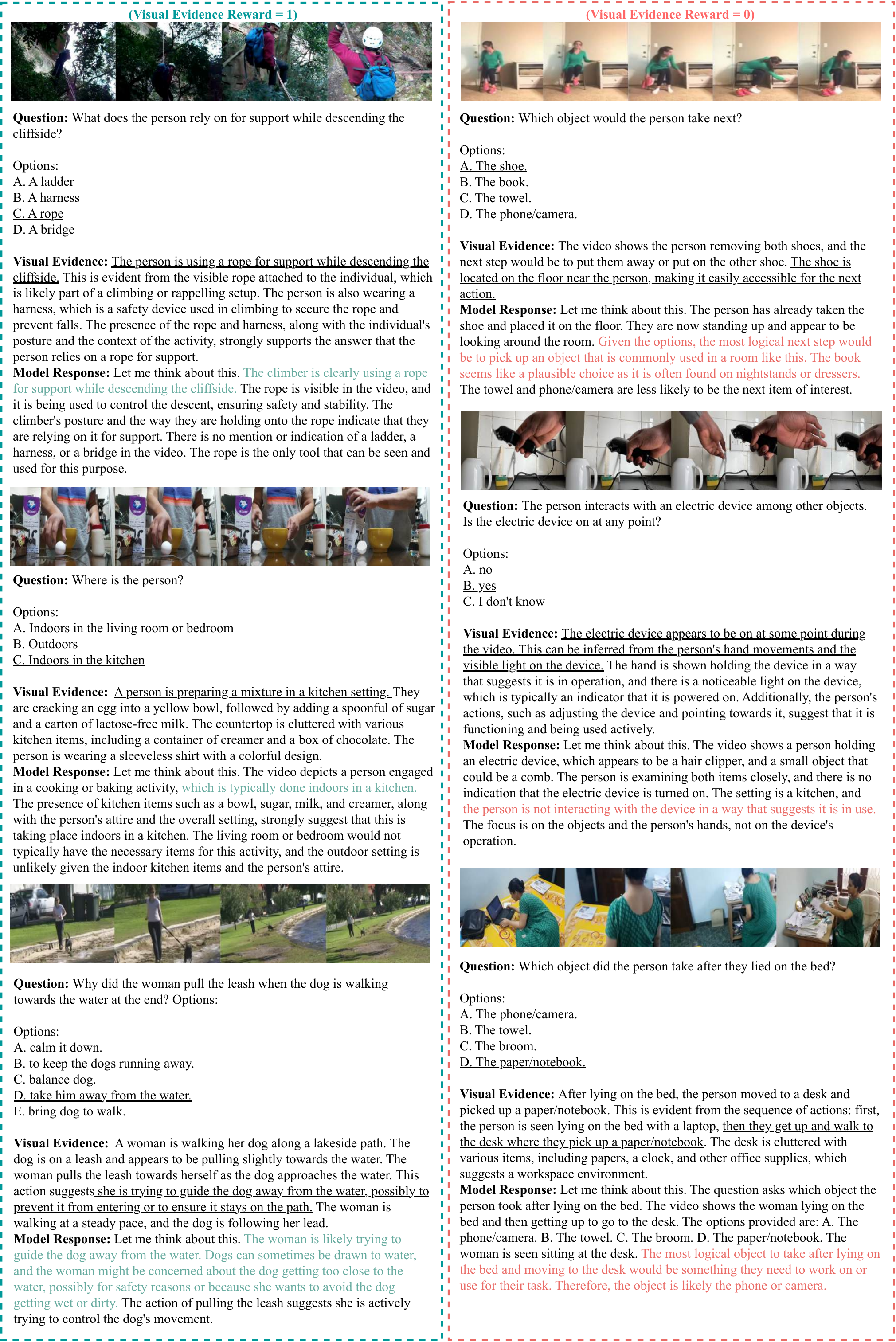}
\caption{Visual evidence generated through our \emph{inverted prompting} approach, as well as the evidence reward values produced by the LLM Judge. These examples illustrate the specific and relevant visual details that directly support answers and anchor the reasoning process. Key visual evidence is underlined. Green highlighting indicates successful reference to visual evidence, while red indicates a failure to do so.}
\label{fig:visual_facts_supp}
\end{figure}

This design aligns with common practice in recent reinforcement learning and reward modeling studies, where LLMs are used to produce reward signals for complex, fuzzy objectives such as factuality, helpfulness, or alignment~\cite{zheng2023judging, gu2024survey, xie2023text2reward}. Unlike token-level metrics, the LLM judge offers flexible, high-level reasoning about textual similarity and visual grounding. While this introduces some approximation and risk of bias, it is essential for enabling scalable supervision at training time where human annotation is impractical. To reduce their influence we use temperature-0 decoding for more deterministic decoding and use the carefully designed prompt shown in Figure~\ref{fig:prompt_llm_judge}. In Figure~\ref{fig:visual_facts_supp}, we present qualitative examples where the LLM-assigned rewards reflect accurate judgments of visual grounding, demonstrating the judge’s effectiveness in identifying whether reasoning is properly anchored in the video content.

\section{Visual Evidence Generation: Bootstrapping Grounded Reasoning}

Generating high-quality, question-specific visual evidence is crucial for anchoring Chain-of-Thought (CoT) reasoning in video understanding. This task demands identifying precise visual cues---objects, actions, and temporal events---that directly justify a given answer. Given the current limitations of most video-language models (MLLMs) in reliably producing such granular evidence without explicit supervision, we employ a strong external MLLM, Qwen2.5-VL-72B, as an \textbf{offline generator} during training. Our goal is to bootstrap a lightweight, yet effective, reward signal that explicitly encourages reasoning traces to be grounded in observable visual content, thereby circumventing the need for expensive human annotation.
\custompar{Potential Concerns}
Introducing an external ``teacher'' MLLM for evidence generation naturally raises valid concerns regarding noise, potential circularity in the training process, and the risk of hallucinated details. We've implemented several strategies to rigorously mitigate these issues.

First, and crucially, the external MLLM is utilized \emph{solely offline} to create question-specific visual evidence for training. Our policy model, \modelname, learns from these generated outputs but operates \emph{entirely independently at inference time}. This ensures there's no direct feedback loop or reliance on the teacher's rationale during deployment, preserving the integrity of our trained model.

Second, the reward signal derived from the generated visual evidence is intentionally \emph{binary}. We don't demand exhaustive or perfectly nuanced evidence; instead, the judge model merely verifies if the generated reasoning trace cites \textit{any} verifiable visual fact relevant to the answer. This design choice significantly attenuates the impact of occasional hallucinations or minor inaccuracies from the teacher model: incorrect or missing evidence simply results in a zero reward, rather than actively pushing the policy towards an erroneous trace. This robustness ensures that the reward signal guides the model toward \emph{grounded reasoning}, rather than penalizing minor deviations.

Third, a core innovation in our visual evidence generation strategy lies in our \emph{inverted prompting approach}. Unlike conventional CoT, where the model simultaneously explores reasoning steps and the final answer, we feed the external MLLM the \emph{(question, ground-truth answer)} pair and instruct it to generate visual evidence that \textit{supports} this already-known answer. This inversion offers two benefits. By fixing the answer upfront, the model's task is narrowed considerably. It no longer has to navigate the full reasoning space; its sole objective is to retrieve the \emph{minimal, verifiable visual facts} that logically lead to the given answer (e.g., ``a red ball enters the basket''). Furthermore, in typical CoT, models can sometimes drift into generic narratives before settling on an answer. Our inverted prompt structure forces each generated evidence snippet to directly explain the already-known answer to the specific question asked. This creates an intrinsic alignment pressure: any statement irrelevant to the question or unsupported by the video becomes effectively useless, and thus discouraged by our binary evidence reward. 
The model is incentivized to produce \emph{highly relevant and visually verifiable facts}, directly addressing the ``Visual Thinking Drift'' problem by ensuring reasoning is tethered to pertinent visual cues. 
A potential concern is that providing the question text might allow the MLLM to generate evidence based on linguistic cues, bypassing direct pixel analysis. However, our empirical results in Table~\ref{tab:ve_type} address this: generating visual evidence conditioned on the question text significantly outperforms using generic video captions. This suggests that the question text's primary contribution is not an increase in hallucination; rather, it provides crucial, task-specific context that guides the MLLM towards more relevant visual features, thereby enhancing overall performance.

Put differently, a standard CoT prompt induces the distribution $p(c_{1:T},a \mid q,\mathbf{v})$, where both the reasoning chain and the answer are uncertain. Our evidence prompt, by contrast, conditions on the correct answer and samples from $p(e_{1:K} \mid q,a,\mathbf{v})$. This constitutes a far lower-entropy target that inherently prioritizes \emph{visual grounding over linguistic priors}. 
This fundamental structural difference in how the evidence is generated explains why our teacher model can reliably produce visual evidence \emph{without} itself needing the very reinforcement signal we are about to learn.

Figure~\ref{fig:visual_facts_supp} provides further samples of our generated visual evidence, demonstrating how our \emph{inverted prompting} approach yields highly relevant and specific visual cues that directly support the answers, effectively anchoring the reasoning process.

\section{More Qualitative Examples}
\begin{figure}[ht!]
\centering
\includegraphics[width=0.93\textwidth]{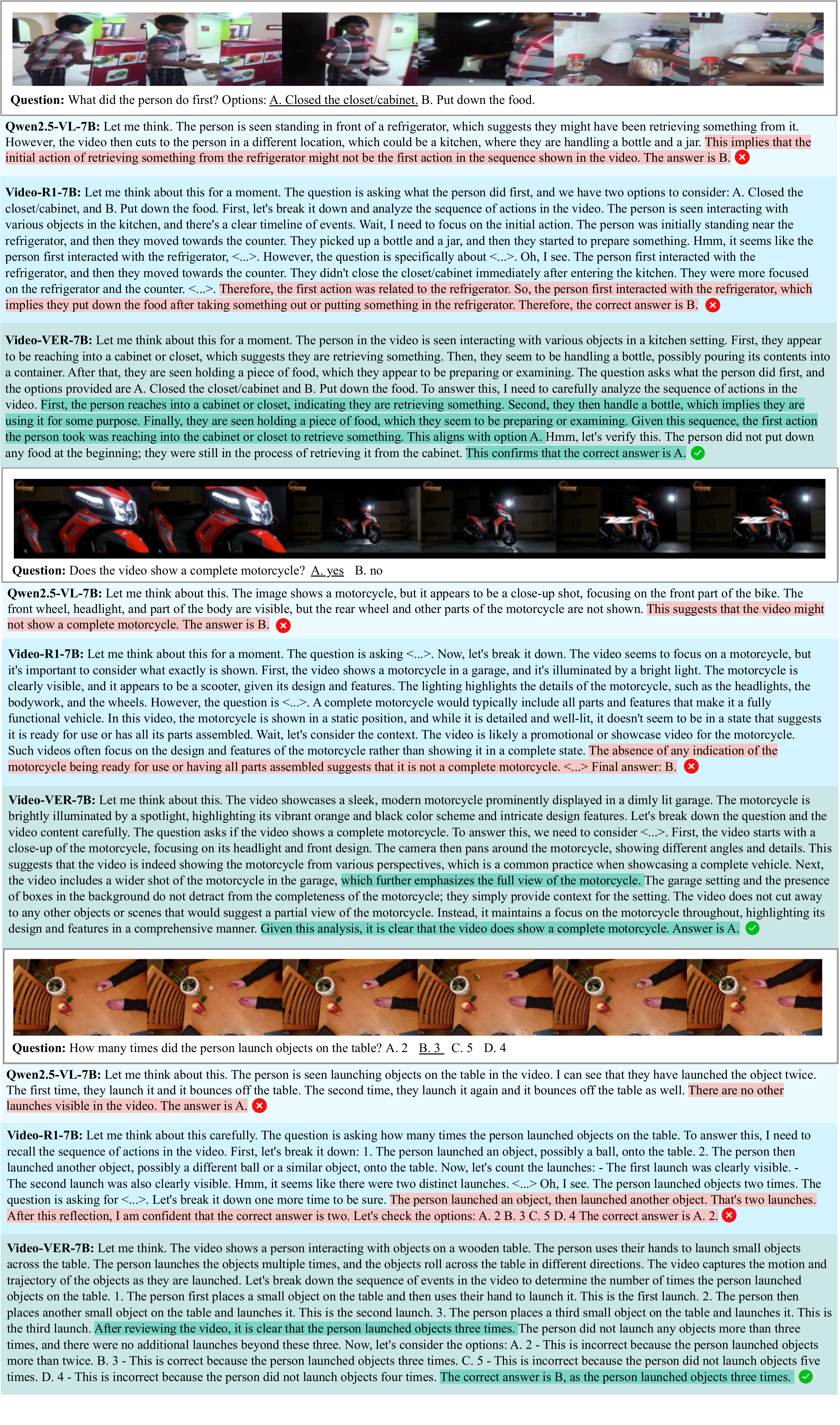}
\vspace{-10pt}
\caption{Comparison of reasoning traces from baseline models and our Video-VER model reveals that baselines often introduce speculative or hallucinated details not grounded in the video. In contrast, Video-VER consistently maintains alignment between reasoning steps and evidence. Text highlighted in red indicates incorrect reasoning steps, while text highlighted in green marks the key logical steps that lead to the correct answer.}
\label{fig:more_quali_1}
\end{figure}

\begin{figure}[ht!]
\centering
\includegraphics[width=0.93\textwidth]{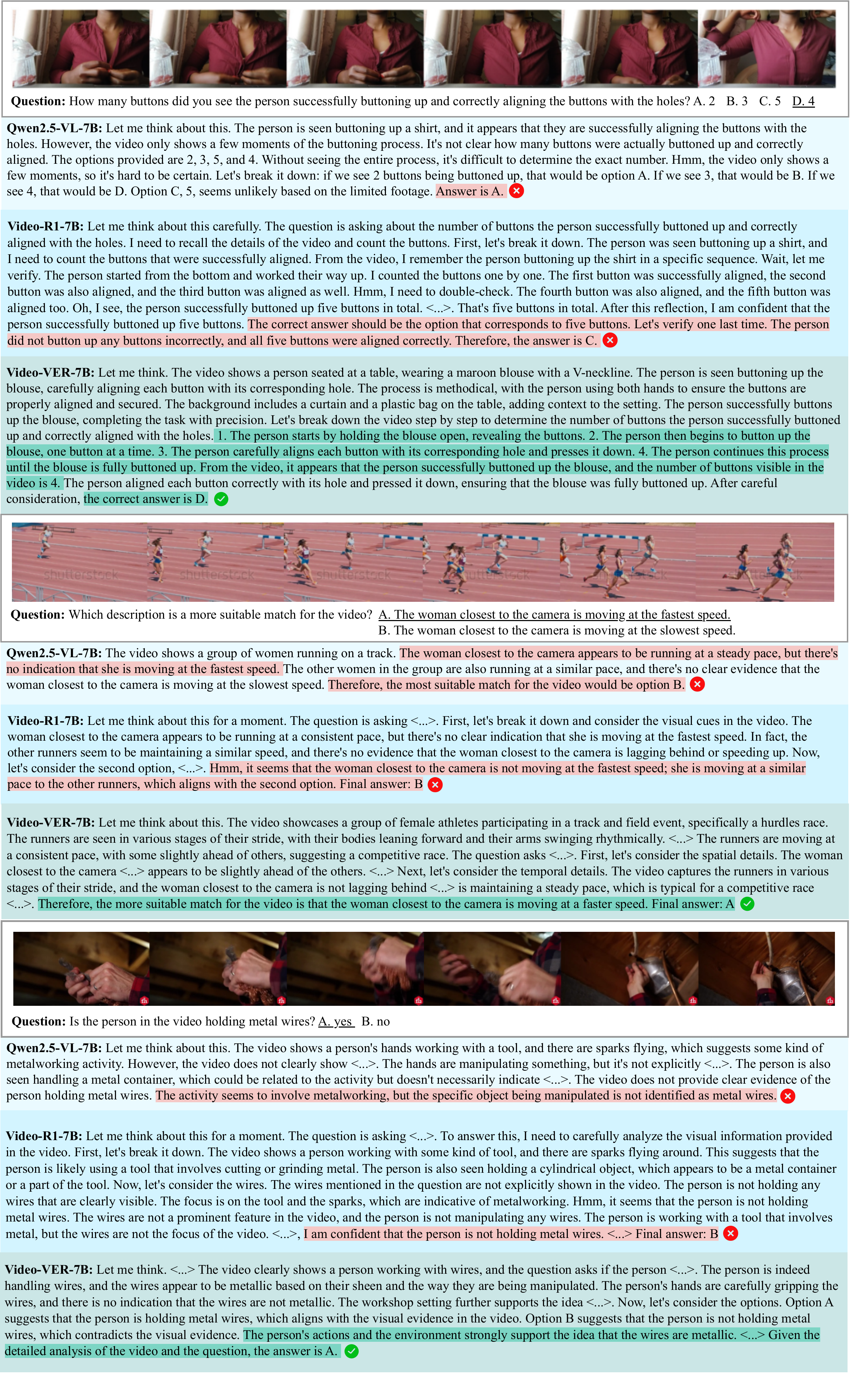}
\vspace{-10pt}
\caption{Comparison of reasoning traces from baseline models and our Video-VER model reveals that baselines often introduce speculative or hallucinated details not grounded in the video. In contrast, Video-VER consistently maintains alignment between reasoning steps and evidence. Text highlighted in red indicates incorrect reasoning steps, while text highlighted in green marks the key logical steps that lead to the correct answer.}
\label{fig:more_quali_2}
\end{figure}

As shown in Figure~\ref{fig:more_quali_1} and~\ref{fig:more_quali_2}, we present additional qualitative examples of reasoning chains produced by our model alongside the baseline models Qwen2.5-VL-7B and Video-R1-7B. These examples further demonstrate that baseline models frequently rely on speculative or hallucinated details, often misaligned with the actual actions or broader temporal context of the video. In contrast, Video-VER consistently grounds its intermediate reasoning steps in observable evidence, resulting in more accurate answers.

\FloatBarrier

\begin{figure}[h!]
\centering
\includegraphics[width=0.99\textwidth]{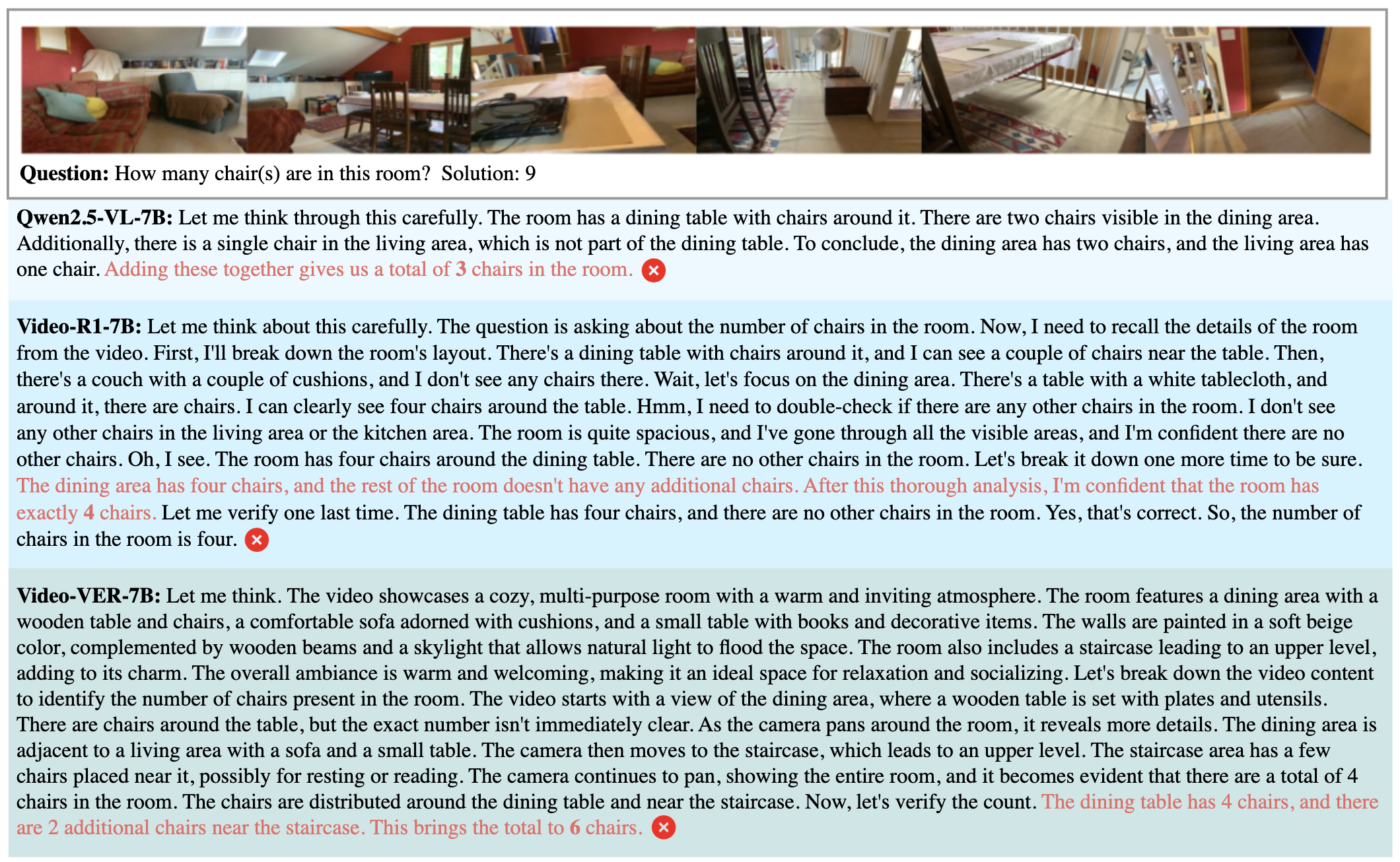}
\caption{A common failure case in which both the baseline models and our method produce incorrect reasoning traces due to missing visual context, caused by incomplete frame sampling.}
\label{fig:failure_case}
\end{figure}

\section{Failure Case}
We illustrate a common failure case where both the baseline models and our Video-VER model generate incorrect reasoning traces. This occurs due to incomplete frame sampling, which omits critical visual context from the video tokens needed to answer the question accurately. See Figure~\ref{fig:failure_case}.

\FloatBarrier

\section{Limitations}

Our \methodname~framework and \modelname~model advance grounded video reasoning. As with any research, there are aspects that provide context for our findings and suggest avenues for future exploration.

First, the performance of video reasoning systems, including \modelname, is closely tied to the nature of the input video and how it is processed. Our current investigations primarily focus on videos of moderate length. This focus considers the current ability of many state-of-the-art vision encoders to effectively process very long video sequences, and the common MLLM design where the language model reasons using visual tokens that are expected to contain all necessary information for the task. As a result, extending robust, fine-grained reasoning to scenarios with very long videos—especially those containing sparse critical information or complex temporal patterns—remains an important direction for future work, which will benefit from advances in long-sequence video encoding. Furthermore, the quality of the visual representation itself significantly influences performance, irrespective of video length. As highlighted in our failure case analysis (Figure~\ref{fig:failure_case}), if essential visual details are missed, for instance, due to frame sampling choices or limitations in feature extraction, the reasoning process can be compromised. Our \methodname~method ensures that reasoning is firmly based on the \textit{available} visual data, and its effectiveness is therefore complemented by ongoing improvements in dynamic frame selection and visual feature extraction techniques.

Second, the LLM-based reward signal generation is a key component of our approach. We utilize Llama-3.1-70B-Instruct with robust prompting strategies and temperature-0 decoding to ensure reliable visual grounding assessment. The quality of the reward signal is inherently connected to the chosen LLM's capabilities. As with similar LLM-based evaluation paradigms in the field, ongoing research into optimizing these judge models and exploring alternative evaluation mechanisms will further enhance the precision and scalability of such reward mechanisms.

Third, regarding the generation of visual evidence for training, our approach employs a powerful external MLLM (Qwen2.5-VL-72B) in an offline manner. Our inventive inverted prompting strategy and binary reward design are specifically configured to maximize the relevance of this evidence and ensure robustness against potential imperfections from the teacher model. While this bootstrapping method proves effective for our training purposes, continued advancements in the capabilities of MLLMs to generate high-fidelity, nuanced visual descriptions will naturally offer opportunities to refine the training data for such reward mechanisms even further.

We view this work as a significant step towards overcoming visual thinking drift and achieving more robust, visually grounded video intelligence. Future directions include extending our \methodname~framework to effectively process long-form videos (potentially incorporating advanced visual information aggregation techniques) and exploring its integration with even more complex multi-step reasoning tasks that demand deeply verifiable and contextually rich thought processes.

\section{Societal Impact}
The advancements in video reasoning presented in this paper, particularly the \methodname~framework designed to mitigate \emph{Visual Thinking Drift} and enhance the grounding capabilities of models like \modelname, offer significant societal potential, encompassing both promising benefits and important ethical considerations. On one hand, more accurate, reliable, and interpretable video understanding systems can yield substantial advantages across diverse domains, including improved accessibility for individuals with visual impairments through richer descriptions of visual media, enhanced educational tools that can better analyze instructional content, and more dependable automated content analysis due to a reduction in model-generated hallucinations.

\end{document}